\documentclass[sigconf]{acmart}

\AtBeginDocument{%
  \providecommand\BibTeX{{%
    \normalfont B\kern-0.5em{\scshape i\kern-0.25em b}\kern-0.8em\TeX}}}

\setcopyright{acmcopyright}
\copyrightyear{2018}
\acmYear{2018}
\acmDOI{XXXXXXX.XXXXXXX}

\acmConference[Conference acronym 'XX]{Make sure to enter the correct
  conference title from your rights confirmation emai}{June 03--05,
  2018}{Woodstock, NY}
\acmPrice{15.00}
\acmISBN{978-1-4503-XXXX-X/18/06}

\acmSubmissionID{3381}

\renewcommand\footnotetextcopyrightpermission[1]{}
\usepackage{multirow}
\usepackage{color}
\begin{document}

\title{A Pre-training Framework for Knowledge Graph Completion}


\author{Kuan Xu}
\authornote{Both authors contributed equally to this research.}
\email{xukuan@bjtu.edu.cn}
\affiliation{%
  \institution{Beijing Jiaotong University}
  \city{Beijing}
  \country{China}
}

\author{Kuo Yang}
\authornotemark[1]
\email{kuoyang@bjtu.edu.cn}
\affiliation{%
  \institution{Beijing Jiaotong University}
  \city{Beijing}
  \country{China}
}

\author{Hanyang Dong}
\email{dong_hanyang@163.com}
\affiliation{%
  \institution{Beijing Jiaotong University}
  \city{Beijing}
  \country{China}
}

\author{Xinyan Wang}
\email{xinyan_wang@bjtu.edu.cn}
\affiliation{%
  \institution{Beijing Jiaotong University}
  \city{Beijing}
  \country{China}
}

\author{Jian Yu}
\email{jianyu@bjtu.edu.cn}
\affiliation{%
  \institution{Beijing Jiaotong University}
  \city{Beijing}
  \country{China}
}

\author{Xuezhong Zhou}
\email{xzzhou@bjtu.edu.cn}
\authornote{Corresponding author}
\affiliation{%
  \institution{Beijing Jiaotong University}
  \city{Beijing}
  \country{China}
}

\renewcommand{\shortauthors}{Kuan Xu, Kuo Yang, Hanyang Dong, Xinyan Wang, and Xuezhong Zhou}

\begin{abstract}
  Knowledge graph completion (KGC) is one of the effective methods to identify new facts in knowledge graph. Except for a few methods based on graph network, most of KGC methods trend to be trained based on independent triples, while are difficult to take a full account of the information of global network connection contained in knowledge network. To address these issues, in this study, we propose a simple and effective \underline{Net}work-based \underline{P}re-training framework for knowl\underline{e}dge gr\underline{a}ph \underline{c}ompl\underline{e}tion (termed NetPeace), which takes into account the information of global network connection and local triple relationships in knowledge graph. Experiments show that in NetPeace framework, multiple KGC models yields consistent and significant improvements on benchmarks (e.g., 36.45\% Hits@1 and 27.40\% MRR improvements for TuckER on FB15k-237), especially dense knowledge graph and higher efficiency (e.g. 6590.83s time consumption for NetPeace and 7967.76s for TuckER on FB15k-237). On the challenging low-resource task, NetPeace that benefits from the global features of KG achieves higher performance (104.03\% MRR and 143.89\% Hit@1 improvements at most) than original models.
\end{abstract}





\maketitle

\section{Introduction}
Knowledge graph (KG) is a semantic network that reveals the relationships between entities, which can formally describe entities and their relationships in the real world. In KG, facts are represented by triples, where each triplet represents a relation between a head entity and a tail entity. For example, a triple $(Paris, capital of, France)$ where $Paris$ and $France$ are called the head and tail entities respectively and $capital of$ is a relation. Lots of KGs, e.g., FreeBase \cite{bollacker2008freebase} and WordNet \cite{miller1995wordnet} have been constructed and applied to many fields, e.g., question answering \cite{huang2019knowledge}, machine translation \cite{zhao2021knowledge} and recommendation systems \cite{guo2020survey}. However, due to the complexity of the real world, these KGs still suffer from the uncertainty and incompleteness. Knowledge graph completion, a.k.a. link prediction (LP) is one of the most promising methods to infer new facts based on known facts. More formally, the goal of KGC is to predict either the head entity in a given query $(?, r, t)$ or the tail entity in a given query $(h, r, ?)$.

In recent years, the researchers have proposed multiple types of KGC methods, including geometric models (e.g., TransE \cite{bordes2013translating}, TransH \cite{wang2014knowledge} and TransR \cite{lin2015learning}), tensor decomposition (TD) models (e.g., TuckER \cite{balavzevic2019tucker}, ComplEx \cite{trouillon2016complex}, DistMult \cite{yang2014embedding} and CP \cite{lacroix2018canonical}), deep learning models (e.g., ConvE \cite{dettmers2018convolutional}, RSN \cite{guo2019learning} and HypER \cite{balavzevic2019hypernetwork}) and reinforcement learning models (e.g., MINERVA \cite{das2017go} and DeepPath \cite{xiong2017deeppath}). As a classical geometric model, TransE \cite{bordes2013translating} regards the relation as a translation from the head entity to the tail entity. Inspired by TransE, many variants e.g., TransH, TransR have been proposed. The key of TD models is to regard the KG as a 3D tensor, then decompose the tensor into a combination of low-dimensional vectors. Besides, other models (e.g., deep leaning \cite{dettmers2018convolutional, guo2019learning, balavzevic2019hypernetwork}, reinforcement learning \cite{xiong2017deeppath, das2017go}) are also applied to LP task. 

Most of KGC methods, such as TuckER \cite{balavzevic2019tucker}, ComplEx \cite{trouillon2016complex}  and DistMult \cite{yang2014embedding} train the model with each fact (triple) as an independent sample.  However, as a special network, there is complex network relationship information between entities in KG. When these models learn the embedding features of entities, they often ignore the complex network relationships between entities, and also ignore the global structural information in the network. Meanwhile, some KGC methods based on graph structure, such as GAKE \cite{feng2016gake}, R-GCN \cite{schlichtkrull2018modeling} do not take the global network connection information into account effectively.

In the field of network embedding (NE), the classical embedding methods, such as DeepWalk \cite{perozzi2014deepwalk}, node2vec\cite{grover2016node2vec}, LINE \cite{tang2015line}  etc., generally learn the latent feature representation of nodes by capturing the relationship among nodes in the entire graph and local or global structure information. The embedding feature that contains network structure information can be directly used for the downstream tasks, e.g., link prediction and node classification. Since NE methods can learn the hidden embedding features of nodes from network structures, why not consider NE to obtain the hidden embedding features of entities from the graph structure in KG, and apply it to the downstream LP tasks by the feature fine-tune of KGC methods.

For the more challenging low-resource task \cite{zhang2020relation} in KGC, previous approaches assume sufficient training triples to learn versatile vectors for entities and relations, or a satisfactory number of labeled sentences to train a competent relation extraction model. However, low resource relations are very common in KGs, and those newly added relations often do not have many known samples for training. As a result, the existing KGC methods do not achieve a satisfactory performance \cite{balavzevic2019tucker, yang2014embedding, trouillon2016complex, dettmers2018convolutional, sun2019rotate}.

Therefore, we proposed NetPeace, a network-based KG pre-training framework that captures global network connections of KG and local triple relationships. NetPeace is applied to downstream KGC tasks and obtains significant improvements on benchmarks. 

Our contributions are as follows:

\begin{itemize}
\item We proposed a network-based KG pre-training framework, which captures global network connections among entities to learn the entity’s embedding features that can be applied to downstream KGC tasks through feature fine-tuning.

\item Comprehensive experiments show that NetPeace obtains significant improvements on benchmarks, especially, obtains better initialized features in denser knowledge graph, and then achieve more significant performance improvement.

\item On the challenging low-resource task, NetPeace that benefits from capturing global features of KG achieves higher performance than original models.
\end{itemize}
\section{RELATED WORK}
Our work is related to knowledge graph completion and network embedding learning. We briefly discuss them in this section.

\subsection{Knowledge Graph Completion}
\paragraph{Geometric Models}
Geometric models interpret the relation as a geometric transformation in the latent space \cite{nguyen2016stranse, bordes2013translating, zhang2019interaction, sun2019rotate}. The distance-based scoring functions are used to measure the plausibility of facts. For example, TransE \cite{bordes2013translating} regards the relation $r$ in each triple $(h, r, t)$ as a translation from the head entity $h$ to the tail entity $t$, by requiring the embedding of tail entity lies close to the sum of the embeddings of head entity and relation. CrossE \cite{zhang2019interaction} learns an additional relation-specific embedding $c_{r}$ for each relation $r$, thereby overcoming the limitation of being a purely transformational model. RotatE \cite{sun2019rotate} represents each relation as a rotation from the head entity to the tail entity in a complex latent space.

\paragraph{Tensor Decomposition Models}
TD models are applied to the LP task in KGs \cite{lacroix2018canonical, yang2014embedding, trouillon2016complex, balavzevic2019tucker}. The key of TD models is to regard the KG as a 3D tensor, then decompose the tensor into a combination of low-dimensional vectors. In CP \cite{lacroix2018canonical} decomposition, a 3D tensor is decomposed into the sum of multiple rank-one 3D tensors. DistMult \cite{yang2014embedding}, a special case of CP decomposition, forces all relation embeddings as diagonal matrices, which reduces the space of parameters and makes the model easier to train. ComplEx \cite{trouillon2016complex} introduces asymmetry into TD by adding complex-valued embeddings, so that it can simulate asymmetric relations. TuckER \cite{balavzevic2019tucker} applies Tucker decomposition to the KGC tasks.

\paragraph{Deep Learning Models}
Deep neural networks are widely used in LP task in KGs \cite{dettmers2018convolutional, schlichtkrull2018modeling, guo2019learning, balavzevic2019hypernetwork}, which usually learn the embeddings of entities and relations using neural network models. For example, ConvE \cite{dettmers2018convolutional}  reshapes head entities and relations into 2D matrices to simulate their interactions, and finally predict candidate entities based on convolutional neural network. RSN \cite{guo2019learning} combines recurrent neural network with residual to learn relation paths, which effectively captures the long-term relational dependencies. HypER \cite{balavzevic2019hypernetwork} is a simplified convolutional model, that uses a hypernetwork to generate 1D convolutional filters for each relation, extracting relation-specific features from subject entity embeddings. Since KG is a kind of graph, some recent works have applied graph convolutional network (GCN) \cite{kipf2016semi} to LP task \cite{schlichtkrull2018modeling}. 

\subsection{Network Embedding}
Network embedding (NE) is an algorithmic framework for learning continuous feature representations for nodes in networks. Classic NE methods, e.g., DeepWalk \cite{perozzi2014deepwalk}, LINE \cite{tang2015line}, SDNE \cite{wang2016structural} and node2vec \cite{grover2016node2vec} usually consider local and global connections between nodes to learn the embedding features of nodes, which can be applied to downstream tasks, e.g., node classification \cite{dalmia2018towards} and link prediction \cite{fu2017hin2vec}. DeepWalk and node2vec used random walk strategy to generate contexts of each node in network, and obtained the nodes’ embedding features, which imply the local structure information of networks. Besides, LINE and SDNE believe that the connected nodes and nodes with common neighbors have higher similarity. Several studies \cite{dalmia2018towards, tang2015line, perozzi2014deepwalk, wang2016structural, grover2016node2vec,fu2017hin2vec} have shown that NE methods can achieve good performance on different downstream tasks. 

\section{PROBLEM FORMULATION}
\paragraph{Definition of Knowledge Graph}
A knowledge graph can be denoted as $\mathcal{G} = \{\mathcal{E}, \mathcal{R}, \mathcal{P}\}$. $\mathcal{E}$ and $\mathcal{R}$ are the entity set and the relation set, respectively. In specific, the relation set $\mathcal{R}$ contains high-frequency relations and low-resource relations. $\mathcal{P} = \{(h, r, t) \in \mathcal{E} \times \mathcal{R} \times \mathcal{E}\}$ denotes the set consisting of all the triple facts in the knowledge graph.

\paragraph{Low-Source Problem of Knowledge Graph}
Most KG embedding models for KGC assume sufficient training triples which are indispensable when learning versatile vectors for entities and relations. However, as illustrated in \cite{zhang2020relation},there is a large portion of relations in KG (e.g., FB15k) with only a few triples, revealing the common existence of low resource relations. And the results of low resource relations are much worse than those of highly frequent ones.

\section{Network-based Pre-training Framework}
In this section, we introduce the Network-based Pre-training Framework (NetPeace). Taking LINE and TuckER as an example, We start by introducing the network embedding e.g., LINE \cite{tang2015line} and tensor decomposition model e.g., TuckER \cite{balavzevic2019tucker}. Then we describe how to
construct the pre-training model based on the two methods mentioned above. In the end, we provide the details of the
training and optimization of the proposed method.

\subsection{Network Embedding Model}
We take the classic NE method, i.e., LINE \cite{tang2015line} as an example to show network embedding. LINE optimizes a carefully designed objective function that preserves both the local and global network structures. LINE can be applied to large networks, which is suitable for arbitrary types of information networks: undirected, directed, and/or weighted. For the embedding vector of each node, it performs low-dimensional embedding of nodes by calculating the first-order  proximity or second-order  proximity between any two nodes.

\paragraph{LINE with First-order Proximity}
The first-order proximity refers to the local pairwise proximity between the nodes in the network. For each pair of nodes linked by an edge $(u,v)$, the weight $w_{uv}$ on that edge represents the first-order proximity between $u$ and $v$. If no edge is observed between $u$ and $v$, their first-order proximity is 0. For each undirected edge $(i,j)$ in the network, it defines the joint probability between nodes $v_i$ and $v_j$ as follows:
\begin{equation}
    p (v_i,v_j )=\frac{1}{1+exp({\vec{u}_{i}}^{T}\cdot{\vec{u}_{j}}^{T} )}
\end{equation}
where ${\vec{u}_{i}}^{T}$, ${\vec{u}_{j}}^{T} \epsilon R^{d}$ represent the low-dimensional vector embedding of node $v_i$, $v_j$ respectively.

\paragraph{LINE with Second-order Proximity}
The second-order proximity between a pair of nodes $(u, v)$ in a network is the similarity between their neighborhood network structures. 
It is not determined by the strength of the connection between two nodes, but by the shared neighborhood structure of the nodes. The general concept of second-order proximity can be interpreted as nodes with shared neighbors are likely to be similar. 
Mathematically, let $p_u = (w_{u,1}, . . . , w_{u,\lvert V \lvert})$ denote the first-order proximity of $u$ with all the other vertices,
then the second-order proximity between $u$ and $v$ is determined by the similarity between $p_u$ and $p_v$. If no vertex is
linked from/to both $u$ and $v$, the second-order proximity
between $u$ and $v$ is 0.


\subsection{Tensor Decomposition Model}
TuckER \cite{balavzevic2019tucker} is a powerful KGC model based on Tucker decomposition. In TuckER, a three-mode tensor $\mathcal{X} \epsilon R^{n_1 \times n_2 \times n_3}$is decomposed into a smaller core tensor $\mathcal{G} \epsilon R^{\gamma _1 \times \gamma_2 \times \gamma_3}$ $(\gamma \ll n)$ and three matrices $U \epsilon R^{n_1 \times \gamma _1}$, $V \epsilon R^{n_2 \times \gamma _2}$, $W \epsilon R^{n_3 \times \gamma _3}$ ($\times _n$ denotes the tensor product along the $n-th$ mode. $U, V, W$ denotes head entities, relations and tail entities respectively). The scoring function for TuckER is defined as (equation \ref{eq:score}), where $(e_{h}, e_{r}, e_{t})$ denotes the embedding features of $(h, r, t)$:
\begin{equation}
    \mathcal{X} = \mathcal{G} \times_1 U \times_2 V \times_3 W
\end{equation}
\begin{equation}\label{eq:score}
    \phi (h, r, t)=\mathcal{G} \times _1 e_{h} \times _2 e_{r} \times _3 e_{t}
\end{equation}

\subsection{Pre-training Framework}
\begin{figure*}[h]
  \centering
  \includegraphics[width=0.8\linewidth]{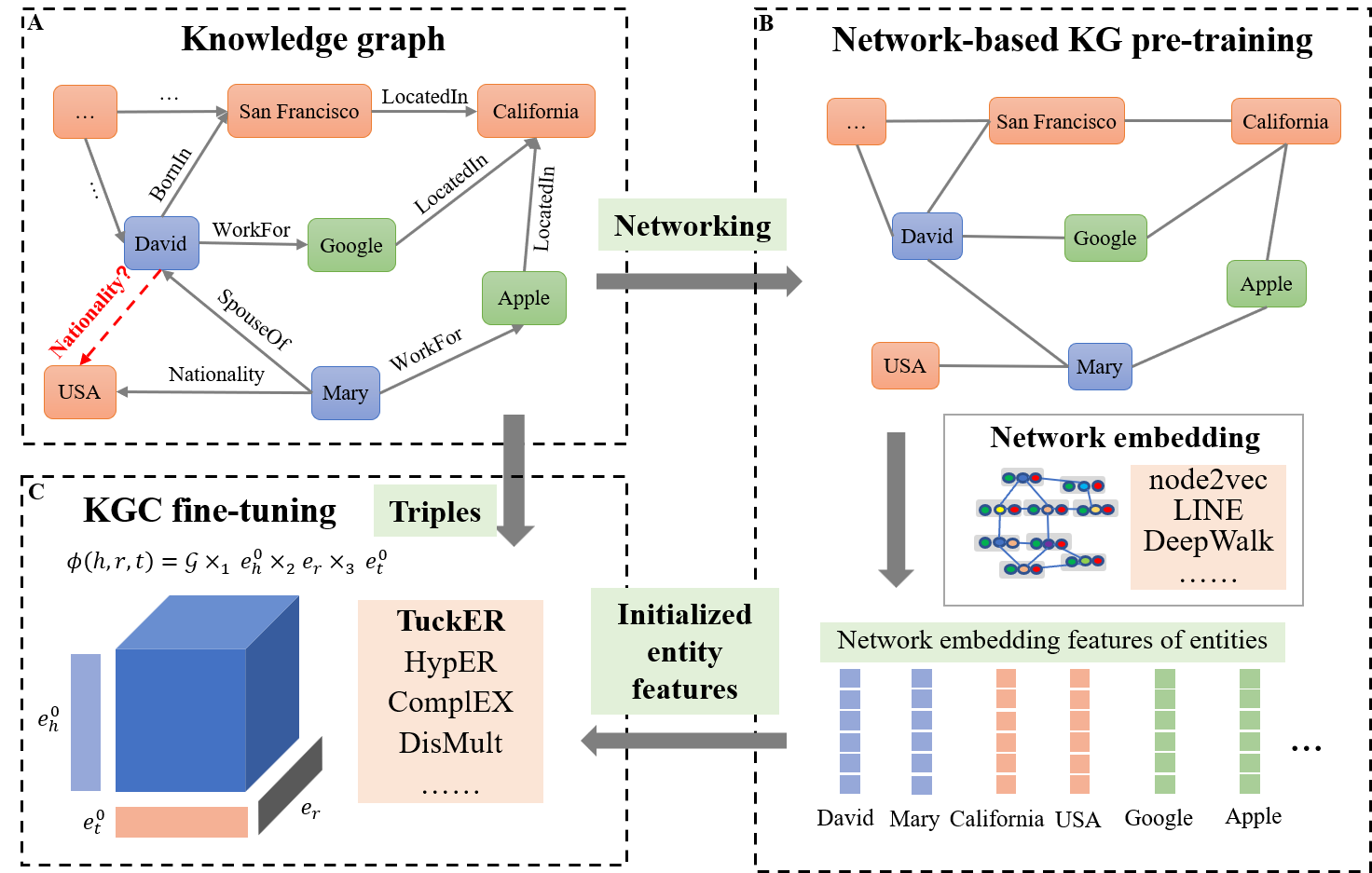}
  \caption{Network-based pre-training framework (NetPeace). (A) An example of knowledge graph. (B) Network-based KG pre-training. (C) Fine-tuning of knowledge graph completion models.}
  \Description{This is about our framework. The figuer consists of three parts. A is about knowledge graph, B is about pre-traing, and C is the knowledge graph completion model.}
  \label{fig:NetPeace}
\end{figure*}

In order to capture the relationship information between entities in the knowledge graph, we propose a network-based KG pre-training framework, which takes the target KG as input graph and uses NE methods to obtain the pre-training features of each entity, and apply it to the downstream KGC task (Figure \ref{fig:NetPeace}).

Mathematically, a triple is represented as $(h, r, t)$, with two entities $h, t \in E$ (the set of entities) and a relation $r \in R$ (the set of relations). We use $e_h, e_t \epsilon R^{d_e}$ to denote the embeddings of head and tail entities and $e_r \epsilon R^{d_r}$ to represent the relation embeddings.

In pre-training framework, the KG are regarded as an undirected and unweighted heterogenous network, which ignores relation attributes in KG. Based on this network, we could utilize network embedding algorithms to learn the embedding features of entities, including head and tail entities, marked by $e_h^0$ and $e_t^0$. These embedding features from pre-training can be used as initial features in following KG completion models. Taking the classic KG completion model, i.e., TuckER, as an example, the scoring function can be defined as:
\begin{equation}
    \phi (h, r, t)=\mathcal{G} \times _1 e_{h}^{0} \times _2 e_{r} \times _3 e_{t}^{0}
\end{equation}

Our proposed pre-training framework is very flexible and simple, and all the proposed network embedding methods can be used to learn and obtain entity embedding features in the pre-training stage. In the later stage of knowledge completion, all currently known KGE methods can use the pre-trained entity embedding features as the initial input, and then improve the performance on the knowledge completion task through model fine-tuning. In fact, each network embedding model has its own characteristics, such as what are the characteristics of LINE and node2vec. A challenging problem is how to choose a suitable network embedding method in the pre-training stage. In this study, we use the experimental performance as the evaluation index, compare several classical network embedding methods, and obtain the pre-trained model and embedding features with the best performance.

\subsection{Training and Optimization}
  The model training phase is mainly divided into two phases. The pre-training stage is network-based entity embedding feature learning, which mainly includes selecting appropriate network embedding methods, embedding feature dimensions, and so on. The fine-tuning stage of the knowledge completion model mainly refers to the respective training strategies of each model. Taking TuckER \cite{balavzevic2019tucker} as an example, we used standard data augmentation techniques \cite{balavzevic2019tucker}, and the same loss function for training (shown in equation \ref{eq:loss}). $p \epsilon R^{n_e }$ represents the vector of predicted probabilities, $y \epsilon R^{n_e }$ is the binary label vector, and $n_e$ represents the number of entities.
\begin{equation} \label{eq:loss}
    L = -\frac{1}{n_e}\sum_{i = 1}^{n_e}(y^{(i)} \log(p^{(i)})+(1- y^{(i)})\log(1-p^{(i)}))
\end{equation}

\section{Experiments}
In this section, we show the details of experiments, including the datasets we used, the experimental setup, and several ablation studies.
\subsection{Datasets}
We use two standard datasets to conduct LP experiments (Table \ref{tab:Datasets}). As a subset of FB15k, FB15k-237 \cite{toutanova2015representing} is built by selecting facts from FB15k incorporating the 401 largest relationships and removing all equivalent or inverse relations. As a subset of WN18, WN18RR \cite{dettmers2018convolutional} contains 11 types of relations and fewer triples than WN18, which filters out all triples with inverse relations. Because of these, FB15k-237 and WN18RR are more challenging.
\begin{table}
  \caption{Summary statistics of datasets}
  \label{tab:Datasets}
  \begin{tabular}{cccccc}
    \toprule
    Dataset & \#Ent & \#Rel & \#Train & \#Dev & \#Test\\
    \midrule
     FB15k-237 & 14,541 & 237 & 272,115 & 17,535 & 20,466\\
     WN18RR & 40,943 & 11 & 86,835 & 3,034 & 3,134\\
  \bottomrule
\end{tabular}
\end{table}

\subsection{Experimental Setup}
During training, we only used the training set provided by the above two datasets for network embedding. We calculated ranks in the filtered setting \cite{bordes2013translating} and used mean reciprocal rank (MRR) and Hits$@$k (k $\in$ \{1, 3, 10\}) as evaluation metrics. MRR is the average of the reciprocal ranks of the true triple over all candidate triples. Hits$@$k is the ratio of predictions for which the rank is equal or lesser than the threshold k.

In order to conduct reliable experiments, We selected several state-of-the-art link prediction methods as baselines and obtained their best published results. Meantime, we reproduced and evaluated HypER and TuckER according to their publicly implementations for fair comparison.

In the setting of experimental hyperparameters, thanks to the work of TuckER \cite{balavzevic2019tucker}, we only adjusted the parameters as follows: relation dimension in \{100, 200, 300, 400\}, entity dimension in \{100, 200, 300, 400\}. The remaining parameters are set according to the optimal parameters of the corresponding data set of TuckER.  At the same time, we tested the first, second-order proximity of LINE \cite{tang2015line}, and the combination of them.

To evaluate the impact of different network embeddings on KGC, we replace the LINE with DeepWalk \cite{perozzi2014deepwalk}, node2vec \cite{grover2016node2vec} and SDNE \cite{wang2016structural}. At the same time, in order to evaluate the impact of network embedding on different models, we fixed LINE and replaced TuckER with ComplEx \cite{trouillon2016complex}, DistMult \cite{yang2014embedding}, ConvE \cite{dettmers2018convolutional}, and HypER \cite{balavzevic2019hypernetwork}. These two parts of experiments are carried out on the FB15k-237 dataset.

To construct datasets for low-resource learning, we go back to original KG (FB15k-237) and select those relations with sparse triples as low-resource relations. We randomly select N (100, 300, 500) samples at most for each relation to form a new training set (For the relations with less triples than N, we select all samples as training set.) and conduct comparative experiments on them.

\section{Results \& Discussion}
\subsection{Overall Results}
We proposed NetPeace is a network-based KG pre-training framework, which could obtain good pre-training features of entities and help KGC models perform better. In the experiments, we took two versions of NetPeace as examples to compare them with several classical KGC methods. The two versions, i.e., NetPeace$_{TK}$ and NetPeace$_{HY}$, conduct KG pre-training based on the same NE method (i.e., LINE) and KGC fine-tuning based on the different methods, i.e., TuckER \cite{balavzevic2019tucker} and HypER \cite{balavzevic2019hypernetwork}, respectively.

\begin{figure}[h]
  \centering
  \includegraphics[width=0.8\linewidth]{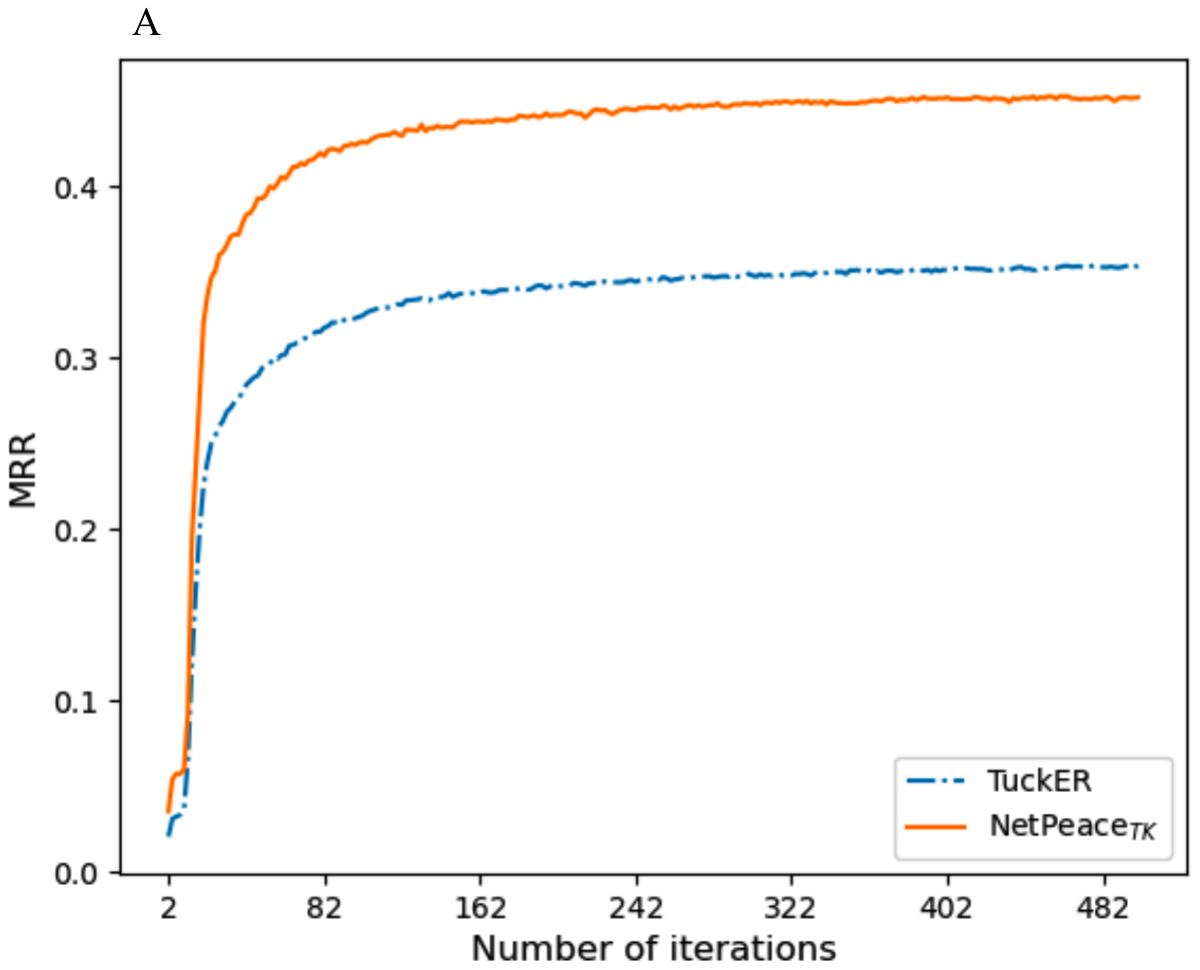}
  \includegraphics[width=0.8\linewidth]{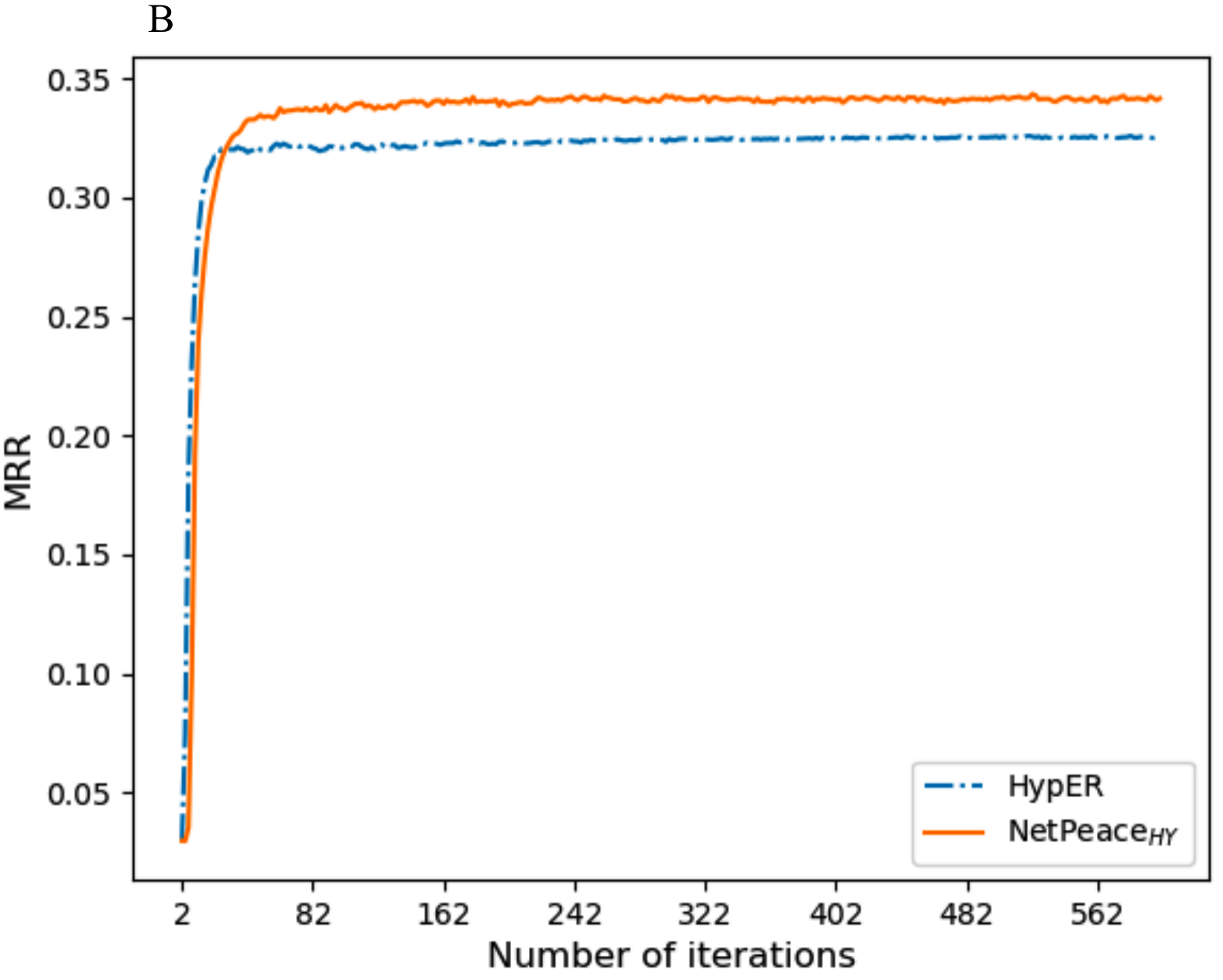}
  \caption{MRR curves on FB15k-237}
  \Description{ In this figure, the curve of NetPeace's MRR is higher than Original Model's with the increasement of number of iterations.}
  \label{fig:Speed}
\end{figure}

Overall, the experimental results (Table \ref{tab:OverallResults}) showed NetPeace obtains comparable performance on all KGC datasets. Compared with all the baselines, NetPeace$_{TK}$ obtain highest performance on FB15k-237 and WN18RR datasets in all the metrics. Especially, compared with the original model TuckER, NetPeace$_{TK}$ has a significant improvement on FB15k-237, including 28.05\% MRR (from 0.353 to 0.452), 36.92\% Hits$@$1 (from 0.26 to 0.356), 27.51\% Hits$@$3 (from 0.389 to 0.496) and 18.52\% Hit$@$10  (from 0.54 to 0.64). Meanwhile, NetPeace$_{HY}$ performs better than the original model HypER. Our experiments also showed that benefiting from fine-tuned initialized features, NetPeace$_{TK}$ and NetPeace$_{HY}$ can achieve higher performance under the same number of iterations (Figure \ref{fig:Speed}). Overall, the above results indicated that In NetPeace framework, the pre-training features of entities does improve the performance of KGC models.

\begin{figure}[h]
  \centering
  \includegraphics[width=0.8\linewidth]{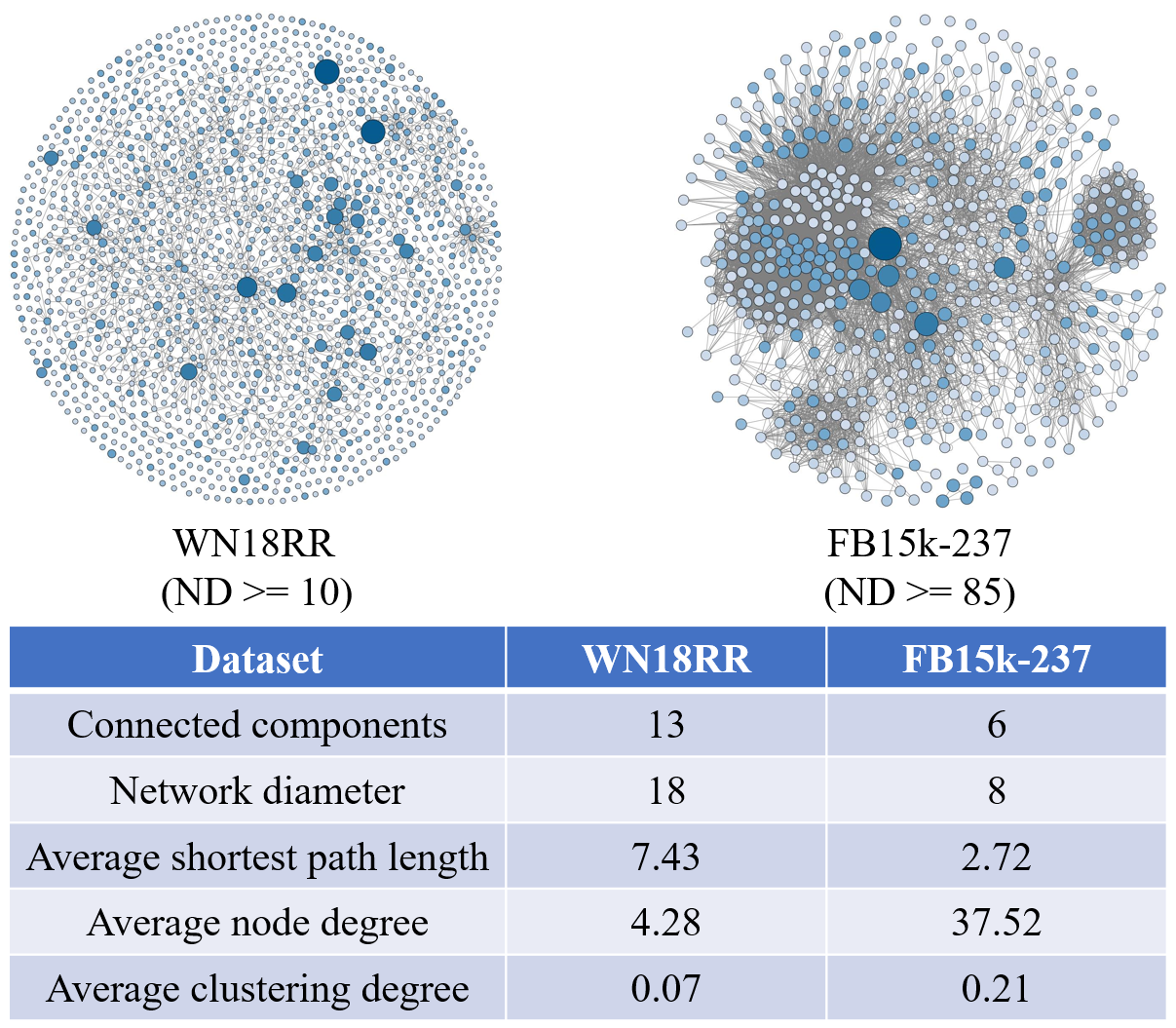}
  \caption{Network visualization and characteristics of WN18RR and FB15k-237 (ND means node degree).}
  \Description{ This picture shows that FB15k-237 is denser than WN18RR.}
  \label{fig:NetFeature}
\end{figure}

To investigate the reason that NetPeace obtain larger improvement on performance on FB15k-237 than WN18RR, we analyzed the difference between WN18RR and FB15k-237 from the perspective of network structure (shown in Figure \ref{fig:NetFeature}), which shows that FB15k-237 is a denser network (i.e., more network connections among nodes) than WN18RR, including fewer connected components than WN18RR (6 vs. 13), shorter network diameter (8 vs. 18), smaller average shortest path length (2.72 vs. 7.43), higher average node degree (37.52 vs. 4.28), higher average clustering degree (0.21 vs. 0.07). Namely, NetPeace has powerful ability to capture the better embedding features of entities from denser knowledge network, finally obtained larger improvement on the performance of KGC tasks.

\begin{table*}
  \caption{Main results on WN18RR and FB15k-237. MRR and Hits@k metrics are filtered. The Reported is from \cite{balavzevic2019tucker}. TK, HY means TuckER and HypER respectively.}
  \label{tab:OverallResults}
  \resizebox{0.8\linewidth}{!}{
  \begin{tabular}{ccccccccccl}
    \toprule
    & \multirow{2}{*}{Model} & \multicolumn{4}{c}{WN18RR}&\multicolumn{4}{c}{FB15k-237}\\
    &                        &MRR&Hits$@$1&Hits$@$3&Hits$@$10&MRR&Hits$@$1&Hits$@$3&Hits$@$10\\
    \midrule
    \multirow{9}{*}{\rotatebox{90}{Reported}}& DistMult \cite{yang2014embedding} &.43&.39&.44&.49&.241&.155&.263&.419\\
                                             &ComplEx \cite{trouillon2016complex}&.44&.41&.46&.51&.247&.158&.275&.428\\
                                             &Neural LP \cite{yang2017differentiable}&-&-&-&-&.25&-&-&.408\\
                                             &R-GCN \cite{schlichtkrull2018modeling}&-&-&-&-&.248&.151&.264&.417\\
                                             &ConvE \cite{dettmers2018convolutional}&.43&.4&.44&.52&.325&.237&.356&.501\\
                                             &M-Walk \cite{shen2018m}&.437&.414&.445&-&-&-&&\\
                                             &RotatE \cite{sun2019rotate}&-&-&-&-&.297&.205&.328&.48\\
                                             &HypER \cite{balavzevic2019hypernetwork}&.465&.436&.477&.522&.341&.252&.376&.52\\
                                             &TuckER \cite{balavzevic2019tucker}&.47&\underline{.443}&\underline{.482}&.526&\underline{.358}&\underline{.266}&\underline{.394}&\underline{.544}\\
    \midrule
    \multirow{4}{*}{\rotatebox{90}{Experiments}}&HypER&.461&.433&.474&.517&.325&.238&.356&.502\\
                                                &TuckER&.462&.434&.473&.515&.353&.26&.389&.54\\
                                                &NetPeace$_{HY}$&\underline{.471}&.441&\underline{.482}&\underline{.528}&.344&.256&.376&.526\\
                                                &NetPeace$_{TK}$& \textbf{.489}&\textbf{.459}&\textbf{.505}&\textbf{.545}&\textbf{.452}&\textbf{.356}&\textbf{.496}&\textbf{.64}\\
  \bottomrule
\end{tabular}
}
\end{table*}

\subsection{KG Pre-training Improving Different KGC Models}
\begin{figure}[h]
  \centering
  \includegraphics[width=\linewidth]{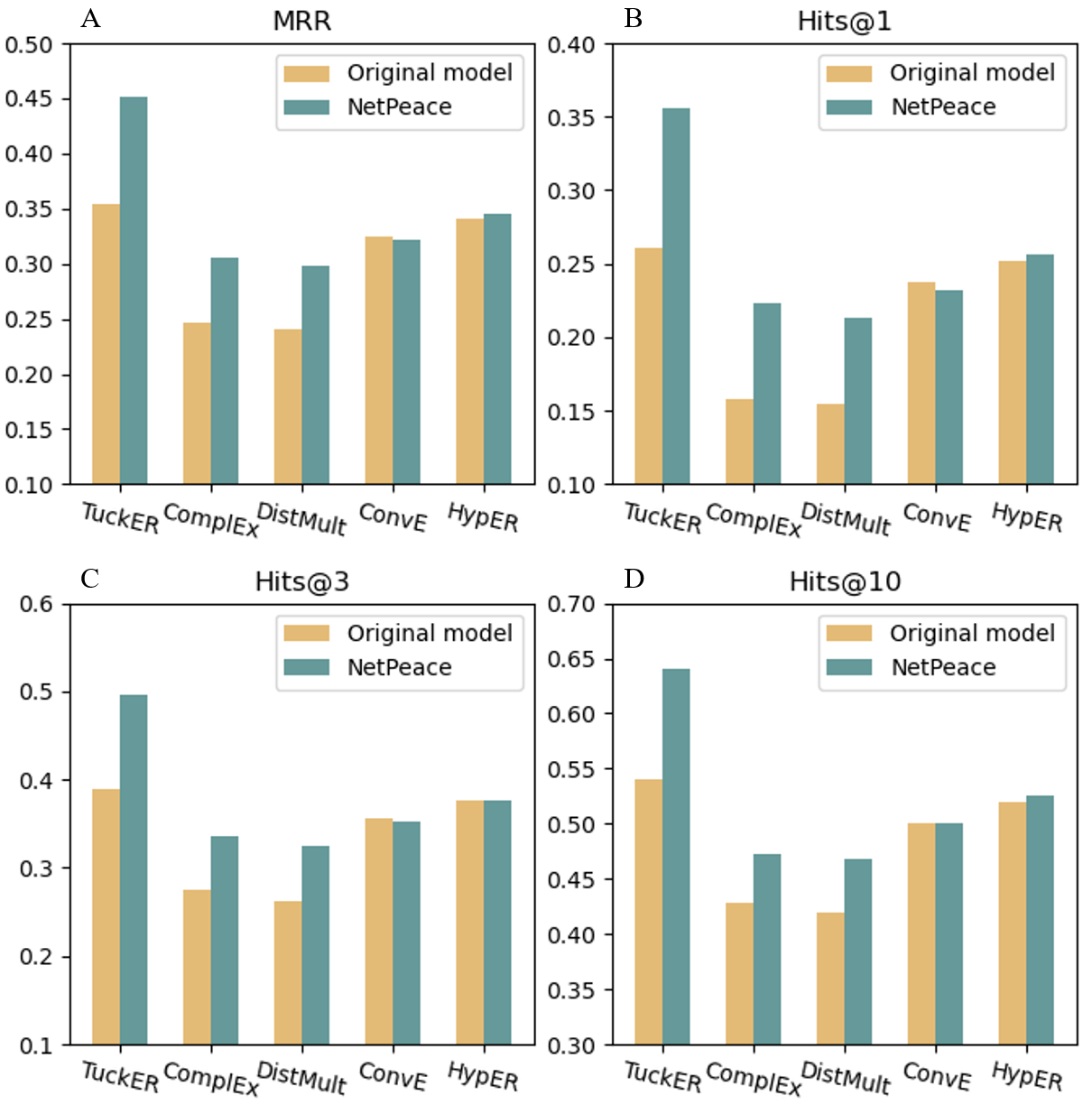}
  \caption{Influence of KG pre-training on different KGC models.}
  \Description{This picture shows that NetPeace's influence on model of TD is higher than neural network. }
  \label{fig:DiffKGC}
\end{figure}
To explore the influence of KG pre-training on different KGC models, we conducted comprehensive experiments, which selected FB15k-237 as benchmark and compared the performance of several KGC models, i.e., TuckER, ComplEx, DistMult, ConvE and HypER in NetPeace framework. The results (Figure \ref{fig:DiffKGC}) showed that there are significantly improvements on the performance of NetPeace with TuckER, ComplEx and DistMult. The performance on NetPeace with ConvE and HypER is basically the same as the original models. According to the types of these KGC models, TuckER, ComplEx and DistMult are based on tensor decomposition, and ConvE and HypER are based on neural networks. That is to say, the KG pre-training of NetPeace does significantly improve the performance of the KGC models based on tensor decomposition, hardly improve the performance of that based on neural networks. The possible reason is that compared with tensor decomposition, the multi-layer neural networks of ConvE and HypER have difficulty to update gradient information to initialized input features, which hardly affects the performance of the KGC models.

\subsection{NetPeace with Different KG Pre-training}
\begin{figure*}[h]
  \centering
  \includegraphics[width=0.8\linewidth]{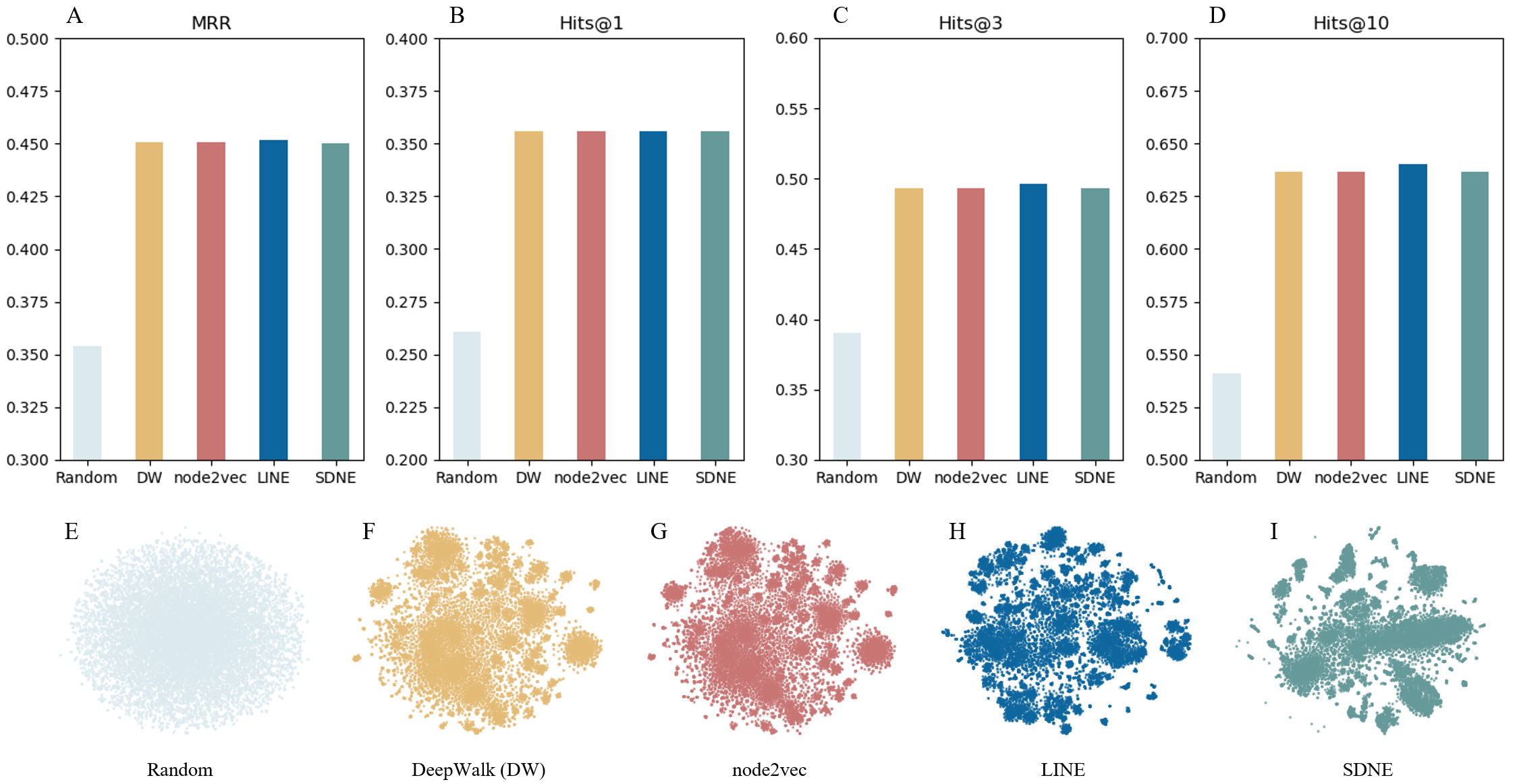}
  \caption{Comparison of different pre-training models in NetPeace. (A-D) The performance comparison of different pre-training models. (E-I) Visualization of the pre-trained features using t-SNE \cite{van2008visualizing}.}
  \Description{This figure shows that the embedding of different pre-traing methods has its own characteristic and is different from random. }
  \label{fig:DiffPre}
\end{figure*}

To explore the influence of different KG pre-training on NetPeace, we took TuckER as the KGC model and compared the performance of five KG pre-training based on network embedding, LINE, node2vec, DeepWalk, SDNE and Random (namely original TuckER) on FB15k-237 dataset. The results (Figure \ref{fig:DiffPre}) showed compared with random initialized features, the four pre-training models all improve the performance of the KGC model. We reduced the entity embedding vectors learned from the three NE models and random initialized vectors to two dimensions and visualized them in plane. The results showed that there are obvious spatial module structures (close distance among some nodes) of the node features compared to random features. Namely, NetPeace learns entity embedding features through network embedding that imply the potential network connections among them, which is beneficial to improve the performance of KGC models.

\subsection{NetPeace’s Improvement on Low-resource KGC Task}
\begin{table*}
  \caption{NetPeace’s performance on low-resource KGC tasks. Train$_N$ denotes that there are at most N triples for each relation extracted from FB15k-237. }
  \label{tab:LowResource}
  \resizebox{0.8\linewidth}{!}{
  \begin{tabular}{ccccccc}
    \toprule
    Train samples&\#Train&Model&MRR&Hits$@$1&Hits$@$3&Hits$@$10\\
    \midrule
    \multirow{2}{*}{Train$_{100}$}&\multirow{2}{*}{23,619 (8.68\%)}&TuckER&0.0967&0.0597&0.1022&0.1748\\
     & &NetPeace&\textbf{0.1973} (\textcolor{red}{${\uparrow}$}104.03\%)&\textbf{0.1456} (\textcolor{red}{${\uparrow}$}143.89\%)&\textbf{0.2119} (\textcolor{red}{${\uparrow}$}107.34\%)&\textbf{0.3029} (\textcolor{red}{${\uparrow}$}73.28\%)\\
    
    \multirow{2}{*}{Train$_{300}$}&\multirow{2}{*}{57,605 (21.17\%)}&TuckER&0.1646&0.1123&0.172&0.2711\\
    & &NetPeace&\textbf{0.2511} (\textcolor{red}{${\uparrow}$}52.55\%)&\textbf{0.1896} (\textcolor{red}{${\uparrow}$}68.83\%)&\textbf{0.2628} (\textcolor{red}{${\uparrow}$}52.79\%)&\textbf{0.3768} (\textcolor{red}{${\uparrow}$}38.98\%)\\
    
    \multirow{2}{*}{Train$_{500}$}&\multirow{2}{*}{80,562 (29.61\%)}&TuckER&0.192&0.1329&0.2031&0.3099\\
    & &NetPeace&\textbf{0.2708} (\textcolor{red}{${\uparrow}$}41.04\%)&\textbf{0.2025} (\textcolor{red}{${\uparrow}$}52.37\%)&\textbf{0.2871} (\textcolor{red}{${\uparrow}$}41.36\%)&\textbf{0.4074} (\textcolor{red}{${\uparrow}$}31.46\%)\\
    
    \multirow{2}{*}{Train$_{all}$}&\multirow{2}{*}{272,115 (100.00\%)}&TuckER&0.353&0.260&0.389&0.540\\
    & &NetPeace&\textbf{0.452} (\textcolor{red}{${\uparrow}$}28.05\%)&\textbf{0.356} (\textcolor{red}{${\uparrow}$}36.92\%)&\textbf{0.496} (\textcolor{red}{${\uparrow}$}27.51\%)&\textbf{0.640} (\textcolor{red}{${\uparrow}$}18.52\%)\\
  \bottomrule
\end{tabular}
}
\end{table*}
A common challenge in KGC tasks is low resource problem, where there lacks of enough triple samples to train an effective KGC model for those newly added relations. In this study, we conducted a low resource experiment, which randomly select N (100, 300 and 500) samples at most for each relation from FB15k-237 dataset to form three new training set (Train$_{100}$, Train$_{300}$, Train$_{500}$), and the test set remains the same. We use NetPeace to conduct NE pre-training for the new training set and compared the performance on KGC tasks. The results (Table \ref{tab:LowResource}) showed that NetPease’s performance significantly outperforms TuckER on all low-resource dataset. For example, for the dataset Train$_{100}$ that only contains 23,619 samples (100 samples at most for each relation, 8.68\% of the full dataset), NetPeace obtained 104.03\%, 143.89\%, 107.34\%, 73.28\% improvements on MRR, Hits$@$1, Hits$@$3 and Hits$@$10 than TuckER, respectively. In particularly, with the decrease of the number of training samples, the improvement is more obvious. This indicates that on the challenging low-resource task, our proposed NetPeace has the ability of improving the performance.

\subsection{Parameter Sensitivity of NetPeace}

\begin{figure*}[h]
  \centering
  \includegraphics[width=0.8\linewidth]{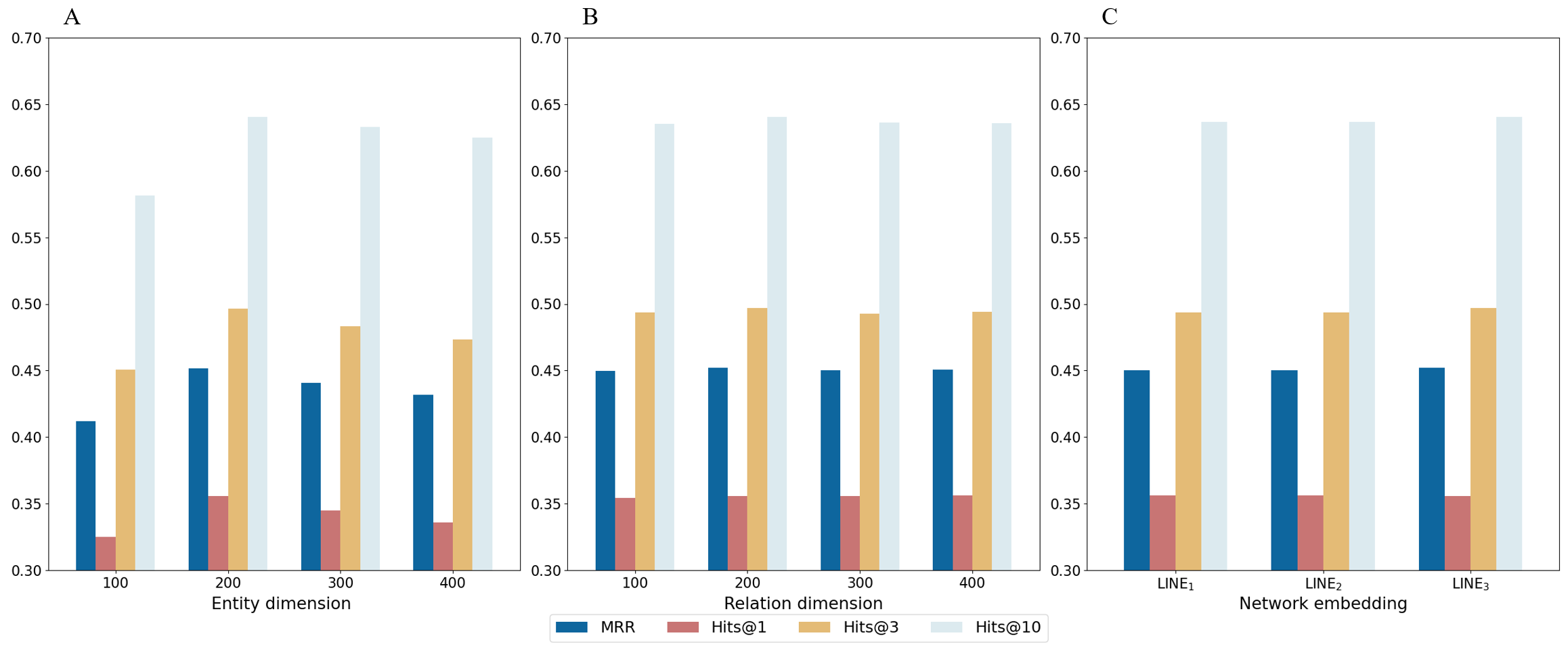}
  \caption{Parameter sensitivity of NetPeace on FB15k-237. In this experiment, NetPeace is based on network embedding model LINE and KGC model TuckER. (A) Entity dimension in \{100, 200, 300, 400\}. (B) Relation dimension in \{100, 200, 300, 400\}. (C) The network embedding model in {LINE$_1$ (the version with first-order proximity), LINE$_2$ (the version with Second-order Proximity), LINE$_3$ (the version with first-order and second-order proximity)}.}
  \Description{Only entity dimension have an influence on NetPeace's performance.}
  \label{fig:Para}
\end{figure*}

To investigate the influence of the hyper-parameters of NetPeace on KGC’s performance, we conducted comprehensive experiments of NetPeace’s parameter tuning on FB15k-237 (Figure \ref{fig:Para}). In the experiment, we tuned three parameters, i.e., entity dimension, relation dimension and network embedding type based on LINE. It shows that the entity dimension has an influence on the performance of NetPeace. Namely, with the increase of entity dimension, the performance tends to increase first and then decrease, and NetPeace obtained the highest performance when the entity dimension was set to 200. In addition, there is no obvious performance difference of relation dimension, network embedding type. Overall, except for entity dimension, other parameters have slight influence on the prediction performance, which imply that NetPeace has some robustness.

\section{Conclusion and future work}
In this study, we propose a new perspective on knowledge complement, which constructed a network-based pre-training framework for improving the performance of downstream KGC by the fine-tuning of features. The experimental results also confirm that NetPeace can do this, i.e., achieves remarkable improvements on benchmarks and low-resource datasets.

There are two potential points for NetPeace in future. On the one hand, our experiments showed neural network based KGC models does not achieve obvious performance improvements in NetPeace framework. In future, we will try to find the real reason for this phenomenon and design a KG pre-training framework that really fits the deep model. On the other hand, we will also explore to utilize network embedding to share knowledge among multi-hop neighbors \cite{das2017go, lin2018multi} to obtain competitive performance on KGC tasks.

\begin{acks}
We want to express gratitude to the anonymous reviewers for their hard work. This work is partially supported by the National Natural Science Foundation of China (82174533 and 82204941).
\end{acks}

\bibliographystyle{ACM-Reference-Format}
\bibliography{NetPeace}

\end{document}